\documentclass[letterpaper]{article} 
\usepackage{pifont}
\usepackage{amsmath}
\usepackage{bm}
\usepackage{graphicx}
\usepackage{amssymb}
\usepackage{multirow}
\usepackage{aaai25}  
\usepackage{times}  
\usepackage{helvet}  
\usepackage{courier}  
\usepackage[hyphens]{url}  
\usepackage{graphicx} 
\usepackage{natbib}  
\usepackage{caption} 
\usepackage{algorithm}
\usepackage{algorithmic}
\usepackage{booktabs}
\usepackage{newfloat}
\usepackage{listings}
\DeclareCaptionStyle{ruled}{labelfont=normalfont,labelsep=colon,strut=off} 
\floatstyle{ruled}
\newfloat{listing}{tb}{lst}{}
\floatname{listing}{Listing}
\title{CNC: Cross-modal Normality Constraint for Unsupervised \\ Multi-class Anomaly Detection}
\author{
   Xiaolei Wang\textsuperscript{\rm 1,2,3}\equalcontrib, Xiaoyang Wang\textsuperscript{\rm 1,2,3}\equalcontrib, Huihui Bai\textsuperscript{\rm 4}, 
   Eng Gee Lim\textsuperscript{\rm 1},  
   Jimin Xiao\textsuperscript{\rm 1}\thanks{Corresponding author.}
}
\affiliations{
    \textsuperscript{\rm 1}Xi’an Jiaotong-Liverpool University\\
    \textsuperscript{\rm 2}University of Liverpool\\
    \textsuperscript{\rm 3}Dinnar Automation Technology\\
    \textsuperscript{\rm 4}Beijing Jiaotong University\\
    \{Xiaolei.Wang, wangxy\}@liverpool.ac.uk, hhbai@bjtu.edu.cn, \{enggee.lim, jimin.xiao\}@xjtlu.edu.cn
    
}

\usepackage{bibentry}

\begin{document}

\maketitle

\begin{abstract}
Existing unsupervised distillation-based methods rely on the differences between encoded and decoded features to locate abnormal regions in test images. However, the decoder trained only on normal samples still reconstructs abnormal patch features well, degrading performance. This issue is particularly pronounced in unsupervised multi-class anomaly detection tasks. We attribute this behavior to `over-generalization' (OG) of decoder: the significantly increasing diversity of patch patterns in multi-class training enhances the model generalization on normal patches, but also inadvertently broadens its generalization to abnormal patches. To mitigate `OG', we propose a novel approach that leverages class-agnostic learnable prompts to capture common textual normality across various visual patterns, and then apply them to guide the decoded features towards a `normal' textual representation, suppressing `over-generalization' of the decoder on abnormal patterns. To further improve performance, we also introduce a gated mixture-of-experts module to specialize in handling diverse patch patterns and reduce mutual interference between them in multi-class training. Our method achieves competitive performance on the MVTec AD and VisA datasets, demonstrating its effectiveness.
\end{abstract}
\begin{links}
\link{Code}{https://github.com/cvddl/CNC}
\end{links}

\section{Introduction}
Visual anomaly detection (AD) mainly focuses on identifying unexpected patterns (deviating from our familiar normal ones) within samples. Industrial defect detection is one of the most widely used branches of AD~\cite{bergmann2019mvtec}, which requires models to automatically recognize various defects on the surface of industrial products, such as scratches, damages, and misplacement. Due to the inability to fully collect and annotate anomalies, unsupervised methods~\cite{yu2021fastflow,gudovskiy2022cflow,liu2023simplenet} become mainstream solutions for AD. Previous unsupervised methods mostly train one model for one class of data, which requires large parameter storage and long training time as the number of classes increases. Therefore, UniAD~\cite{you2022unified} proposed a challenging multi-class AD setting, i.e., training one model to detect anomalies from multiple categories.

\begin{figure*}[t]
\centering
\includegraphics[width=\textwidth]{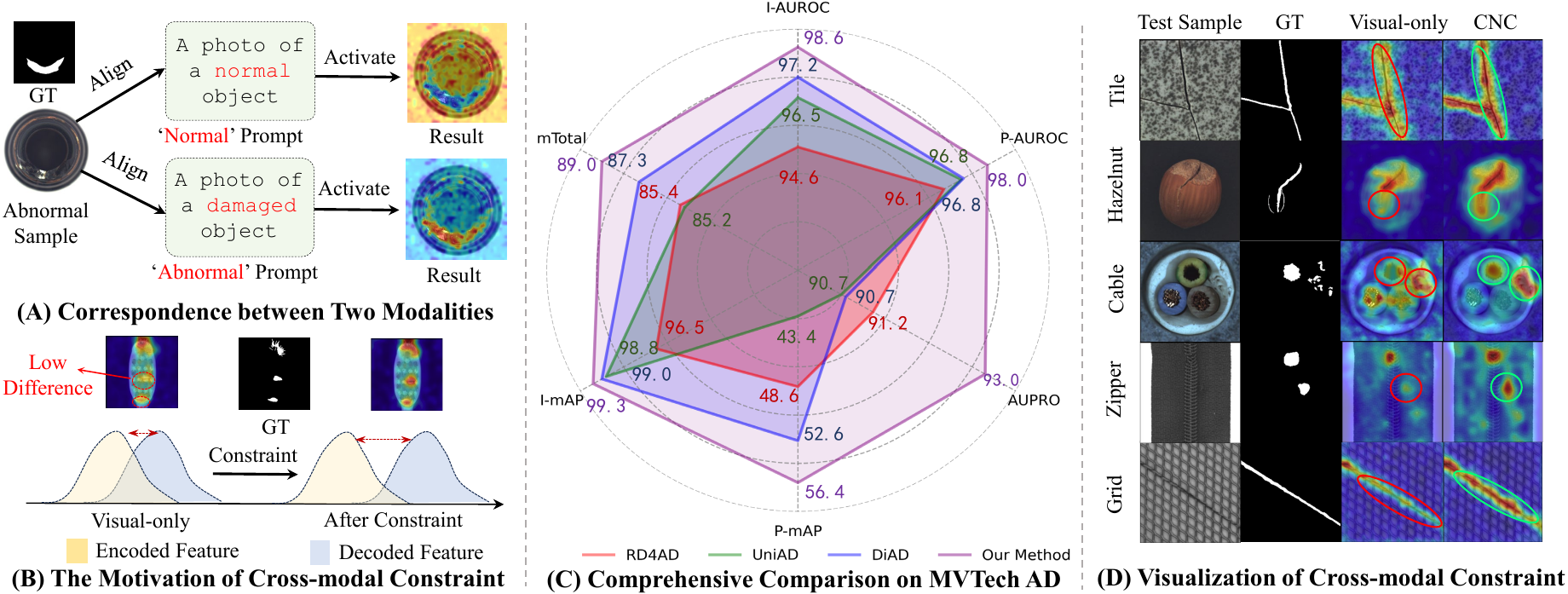}
\caption{(A) and (B) show the correspondence between visual and text modality and the motivation of CNC, respectively. (C) shows a comprehensive performance comparison with previous SOTA methods that only learn sample normality on the visual modality on MVTec AD dataset. (D) gives some visualization results of cross-modal constraint on MVTec AD dataset.}
\label{fig: introduction}
\end{figure*}

Reverse distillation (RD)~\cite{deng2022anomaly} is a highly effective unsupervised AD (UAD) method. It employs a learnable decoder (student network) to reconstruct features from a pre-trained encoder (teacher network) on normal samples via patch-level cosine distance minimization. Ideally, the learned decoder should only recover the encoded normal patches, while failing to reconstruct unseen abnormal patterns. Anomaly regions are then detected by comparing features before and after decoding. In multi-class training, a single model is optimized on the normal samples from multiple classes to achieve unified detection. While the increasing training diversity can improve model generalization on reconstructing normal patches, it also leads to undesired generalization to unseen abnormal patch patterns. Consequently, abnormal regions are recovered well during inference, narrowing the difference between encoded and decoded features and degrading detection performance (Fig.~\ref{fig: introduction}(B) I.). We term this issue `over-generalization' (OG). The key question remains: \textit{How can we effectively mitigate `OG' while preserving the generalization on normal samples in the multi-class distillation framework?}

To address this challenge, we seek to incorporate an additional constraint in the decoding process. Leveraging insights from vision-language models (VLMs)~\cite{radford2021learning}, we observe that normal and abnormal regions within a sample exhibit distinct responses to the same text description, as illustrated in Fig.~\ref{fig: introduction}(A). We propose to exploit this distinction in cross-modal response to differentiate the decoding of normal and abnormal patch features, thereby hindering the recovery of abnormal patterns. Specifically, we employ class-agnostic learnable prompts to extract the common normality from encoded visual features across different classes. These prompts serve as anchors in the textual space, aligning the decoded normal features with a universal representation of normality and suppressing the `over-generalization' of the decoder towards abnormal patterns (Fig.~\ref{fig: introduction}(B) II.). We also design a normality promotion mechanism for feature distillation, introducing cross-modal activation as a control coefficient on visual features to increase sensitivity to unexpected abnormal patterns. We term the combination of these two strategies Cross-modal Normality Constraint (CNC), which aims to mitigate the `OG' issue and enhance anomaly localization in multi-class distillation frameworks (see Fig.~\ref{fig: introduction}(D) for visualization results).

Another angle to tackle the `OG' issue is to mitigate the mutual interference among different patch patterns produced by increasing categories during feature distillation learning. The success of previous one-model-one-class settings can be attributed to the separate learning of patterns for each class, without interference from patch patterns from other classes. However, in multi-class training, inter-class interference is unavoidable. To address this issue, we propose constructing multiple expert networks to specialize in handling different patch patterns. We find that the mixture-of-experts (MoE) framework~\cite{shazeer2017outrageously,ma2018modeling} can selectively process distinct patch patterns, assigning each patch a distinct weighted combination of experts to alleviate the mutual interference. By combining a vanilla RD framework with our CNC and MoE, we achieve performances that surpass previous methods~\cite{deng2022anomaly,you2022unified,he2024diffusion} across multiple metrics (see Fig.~\ref{fig: introduction}(C)).

Our work primarily addresses the inherent `over-generalization' issue for distillation frameworks in multi-class training. To this end, we propose two key strategies: cross-modal normality constraint to facilitate visual decoding, and a mixture-of-experts (MoE) module to process diverse patch patterns selectively. We conduct comprehensive experiments to demonstrate the efficacy of our approach, yielding a notable performance gain over single-modal methods. The main contributions of this work are summarized as follows:  
\begin{itemize}  
\item We identify the `OG' issue in multi-class distillation frameworks and propose a two-pronged solution to address this challenge.  
\item We introduce a cross-modal normality constraint to guide visual decoding, effectively reducing the effect of `OG'.  
\item We design a gated MoE module to selectively handle various patch patterns, mitigating inter-pattern interference and enhancing detection performance.  
\item Our novel cross-modal distillation framework, built from scratch, achieves competitive performance on the MVTec AD and VisA datasets.  
\end{itemize}

\section{Related Work}\label{sec:relatedwork}

\begin{figure*}[t]
\centering
\includegraphics[width=\textwidth]{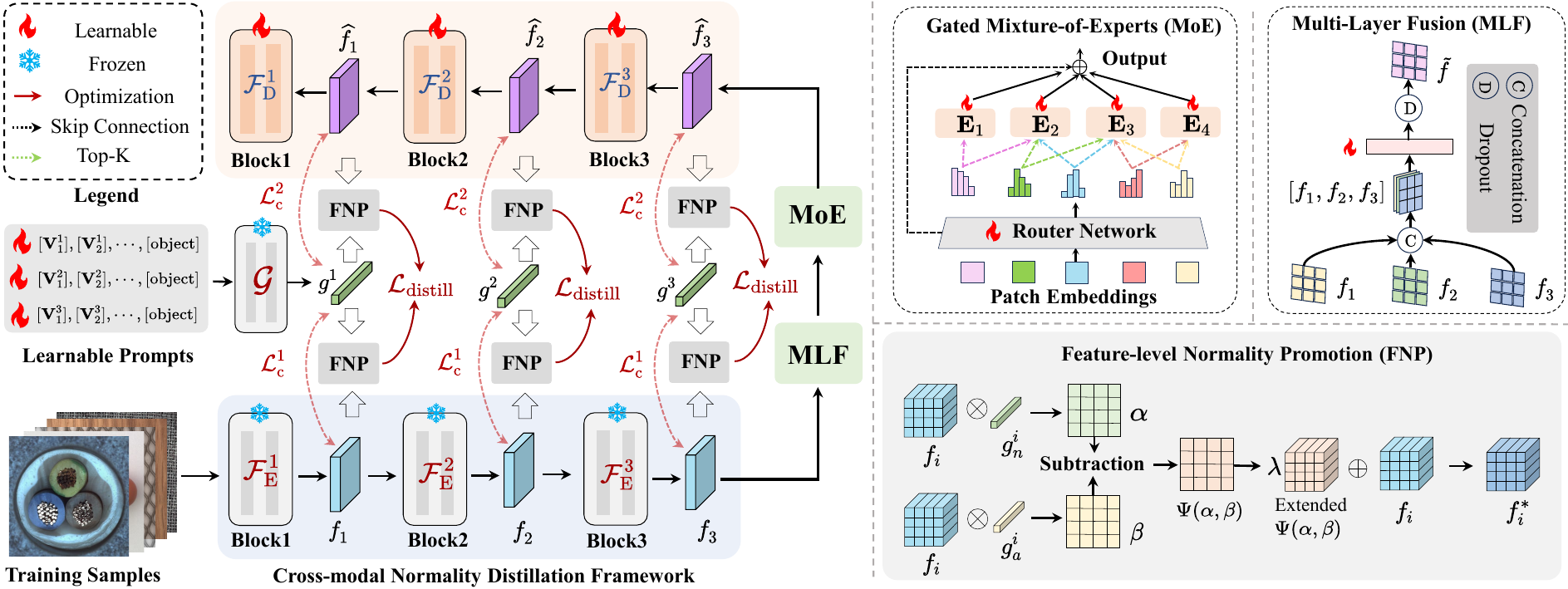}
\caption{Overview of the proposed cross-modal normality distillation framework. Additionally, details of feature-level normality promotion, multi-layer fusion, and gated mixture-of-experts are illustrated in this graph. }
\label{Methodology}
\end{figure*}

\subsection{Anomaly Detection}
Visual anomaly detection contains various settings according to specific engineering requirements, e.g., unsupervised AD~\cite{yi2020patch,zou2022spot,gu2023remembering,cao2023anomaly,liu2023diversity,zhang2024realnet}, zero and few-shot AD~\cite{huang2022registration,fang2023fastrecon,lee2024text}, noisy AD~\cite{chen2022deep,jiang2022softpatch}, and 3D AD~\cite{gu2024rethinking,costanzino2024multimodal,liu2024real3d,li2024towards}. Existing unsupervised AD methods can be roughly divided into reconstruction-based~\cite{tien2023revisiting,lu2023hierarchical,he2024diffusion}, feature-embedding-based~\cite{mcintosh2023inter,roth2022towards,lei2023pyramidflow}, augmentation-based~\cite{zavrtanik2021draem,zhang2024realnet,lin2024comprehensive} methods.
\subsubsection{Reconstruction-based Method}
The reconstruction-based method employs the autoencoder framework to learn the normality of training samples by reconstructing data or its features. Therefore, some works~\cite{zhang2023exploring,guo2024recontrast} rethink RD as reconstruction-based method. Although this type of approach offers fast inference speed, the anomaly localization is inevitably degraded by `OG', which is attributed to the increasing diversity of patch patterns in multi-class training. To address the issue, we propose CNC and MoE to alleviate undesired generalization. 

\subsubsection{Feature-embedding-based Method}
These methods typically rely on pre-trained networks to extract feature embedding vectors from a feature extractor trained on natural images, then apply density estimation~\cite{defard2021padim,yao2023explicit}, memory bank~\cite{bae2023pni}, etc., to detect anomalies. However, there is a significant gap between industrial and natural data, and the extracted embedding may not be suitable for anomaly detection tasks.

\subsubsection{Augmentation-based Method}
Early augmentation methods~\cite{li2021cutpaste,lu2023removing} typically rely on simple handcraft augmentation techniques such as rotation, translation, and texture pasting to generate pseudo samples for discrimination training. However, a huge gap exists between the obtained synthesized samples and the real defect samples. Therefore, some works based on the generative model are proposed recently, such as VAE-based~\cite{lee2024text}, GAN-based~\cite{wang2023defect}, diffusion-based~\cite{hu2024anomalydiffusion,zhang2024realnet} methods. Due to the incomplete collection of defect shapes and categories, it is impossible to synthesize all types of anomalies.

\section{Preliminaries}
\subsubsection{CLIP}
Contrastive Language Image Pre-training~\cite{radford2021learning} (CLIP) is a large-scale vision-language model famous for its multi-modal alignment ability via training with a lot of image-prompt pair data. Specifically, given an unknown image $\mathbf{x}$ and text-prompts $\{\mathbf{p}_{1},\mathbf{p}_{2},\cdots,\mathbf{p}_{J}\}$, CLIP can predict the probability of alignment between $\mathbf{x}$ and every prompt $\mathbf{p}_{j}$ as follows:

\begin{equation}\label{original clip}
p(\mathbf{y}|\mathbf{x})=\frac{\exp(\mathcal{F}(\mathbf{x})\cdot\mathcal{G}(\mathbf{p}_{\mathbf{y} })/\tau)}{\sum^{J}_{j=1}\exp(\mathcal{F}(\mathbf{x})\cdot\mathcal{G}(\mathbf{p}_{j})/\tau)}, 
\end{equation}
where $\mathcal{F}(\cdot)$ and $\mathcal{G}(\cdot)$ are CLIP visual and text encoder, respectively, and $\tau$ is a temperature hyperparameter. Previous work~\cite{jeong2023winclip} adopts handcraft prompts, such as \texttt{a photo of normal/damaged [class]}, to achieve zero-shot anomaly detection.

\subsubsection{Prompt Learning}
Prompt learning~\cite{jia2022visual,zhou2022learning} focuses on optimizing the input prompts to enhance the language or multi-modal model performance on specific tasks. CoOp~\cite{zhou2022learning} introduces a learnable prompt, $\mathbf{p}=[\mathbf{V}_{1}],[\mathbf{V}_{2}],\cdots,[\mathbf{V}_{\mathrm{M}}],[\texttt{class}]$, to achieve few-shot classification, where each $[\mathbf{V}_{m}]$ is a learnable token, and $\mathrm{M}$ is the number of tokens. However, in multi-class UAD task, we do not expect to utilize any class information. Following works~\cite{zhou2023anomalyclip,li2024promptad}, our applied learnable prompts are defined as:
\begin{equation}\label{normal prompt}
    \mathbf{p}_{n}=[\mathbf{V}^{n}_{1}],[\mathbf{V}^{n}_{2}],\cdots,[\mathbf{V}^{n}_{\mathrm{M}}],[\texttt{object}],
\end{equation}
\begin{equation}\label{abnormal prompt}
    \mathbf{p}_{a}=[\mathbf{V}^{a}_{1}],[\mathbf{V}^{a}_{2}],\cdots,[\mathbf{V}^{a}_{\mathrm{M}}],[\texttt{damaged}],[\texttt{object}],
\end{equation}
where $\mathbf{p}_{n}$ and $\mathbf{p}_{a}$ are normal and abnormal prompts respectively. We argue that utilizing category-agnostic prompts enables the learning of common normality patterns across samples from different classes. Different from previous methods, we apply these prompts to learn the normality of training samples and leverage them to guide visual decoding.
\section{Methodology}\label{sec:methodology}

\subsection{Overview}
The framework of the proposed method is shown in Fig.~\ref{Methodology}. Our method consists of three main sections: (1) Visual Distillation Framework; (2) Cross-modal Normality Constraint; (3) Gated Mixture-of-Experts. In (1), we introduce our basic reverse distillation network. In (2), we propose cross-modal normality constraint to ensure decoded features to align with a textual `normal' representation. Additionally, we propose a cross-modal control coefficient on the visual feature to improve sensitivity to abnormal patch patterns, which is called feature-level normality promotion.
In (3), a gated mixture-of-experts (MoE) module is shown in detail to specifically handle various patch patterns. Finally, the inference phase of our method is given for convenience.
 
\subsection{Visual Distillation Framework}

Compared with previous single-modal distillation-based methods~\cite{deng2022anomaly,tien2023revisiting}, we select multi-modal backbone, CLIP-ViT, as encoder, which means that text-modal information can be adopted to improve detection performance. Specifically,
for a given image $\mathbf{x}\in \mathbb{R}^{H_{0}\times W_{0}\times3}$, the CLIP visual encoder $\mathcal{F}$ encodes $\mathbf{x}$ into multi-layer latent space features as $\{f_{i}\}^{N}_{i=1}$, where $f_{i}\in \mathbb{R}^{H\times W\times C}$ represents $i$-th layer feature (feature size in each layer of ViT is consistent), and $N$ is the number of residual attention layers in ViT~\cite{dosovitskiy2020image}. We select $i_{1}$-th, $i_{2}$-th, $i_{3}$-th layer features as visual encoded features. For convenience, let $\mathcal{F}^{1}_{\mathrm{E}}$, $\mathcal{F}^{2}_{\mathrm{E}}$, $\mathcal{F}^{3}_{\mathrm{E}}$ denote $i_{1}$-th, $i_{2}$-th, $i_{3}$-th blocks respectively, and $f_{1},f_{2},f_{3}$ is the corresponding encoded features. For visual decoder, we adopt three general residual attention blocks (the basic module of ViT) together as decoder and extract corresponding features to reconstruct. Therefore, we denote decoder as $\mathcal{F}_{\mathrm{D}}$ with three blocks $\{\mathcal{F}^{i}_{\mathrm{D}}\}^{3}_{i=1}$, with corresponding decoded features as $\{\widehat{f_{i}}\}^{3}_{i=1}$ (see Fig.~\ref{Methodology}). In addition, to further alleviate the `over-generalization', Gaussian noise is applied on the encoded feature $f_{i}$ to obtain its perturbed version $f^{\mathrm{noise}}_{i}$.


\subsection{Cross-modal Normality  Constraint}
To alleviate the unexpected `over-generalization' in multi-class training, we propose a text-modal normality constraint to guide the decoded features towards a ‘normal’ textual representation, suppressing the `over-generalization' of the decoder towards the abnormal direction. The key to our proposed cross-modal normality constraint lies in applying learnable category-agnostic prompts to learn common normality from various normal samples and maintain the semantic consistency of visual encoded and decoded features in textual space during the training phase.

\subsubsection{Learning Cross-modal Normality}
In this section, we aim to apply class-agnostic learnable prompts to learn textural normality from various encoded features. Specifically, for a given image $\mathbf{x}$, we apply $\{\mathcal{F}^{i}_{\mathrm{E}}\}^{3}_{i=1}$ blocks in encoder to obtain multiple layer features $\{f_{i}\}^{3}_{i=1}$. According to Eq.~\eqref{normal prompt} and Eq.~\eqref{abnormal prompt}, we initialize three sets of learnable prompts $\{[\mathbf{p}^{i}_{n},\mathbf{p}^{i}_{a}]\}^{3}_{i=1}$, where $\{\mathbf{p}^{i}_{n}\}^{3}_{i=1}$ are applied to learn textual normality from different layer visual encoded features. 
Then, each prompt pair $[\mathbf{p}^{i}_{n},\mathbf{p}^{i}_{a}]$ is input to CLIP text encoder $\mathcal{G}(\cdot)$, producing the corresponding text feature $[g^{i}_{n},g^{i}_{a}]$, where $g^{i}_{n}=\mathcal{G}(\mathbf{p}^{i}_{n})$, $g^{i}_{a}=\mathcal{G}(\mathbf{p}^{i}_{a})$. Next, we employ a modal-alignment optimization object to learn the textual normality from $\{f_{i}\}^{3}_{i=1}$:
\begin{equation}\label{constrain 1}
    \mathcal{L}^{1}_{\mathrm{c}}=\sum_{i=1}^{3} -\mathrm{log}\frac{\mathrm{exp}(\mathbf{e}_{i}\cdot g^{i}_{n}/\tau)}{\mathrm{exp}(\mathbf{e}_{i}\cdot g^{i}_{n}/\tau)+\mathrm{exp}(\mathbf{e}_{i}\cdot g^{i}_{a}/\tau)},
\end{equation}
where $\mathbf{e}_{i}$ represents the global feature of $f_{i}$, $\tau$ is a temperature coefficient. Next, textual features $\{[g^{i}_{n},g^{i}_{a}]\}^{3}_{i=1}$ are adopted in feature distillation and decoding to alleviate unexpected `over-generalization'.

\subsubsection{Feature Distillation with Normality Promotion}
In this section, textual features $\{[g^{i}_{n},g^{i}_{a}]\}^{3}_{i=1}$ are applied on distillation to improve sensitivity to anomaly patch patterns. Specifically, we first define a cross-modal control coefficient on visual encoded and decoded features $f_{i}$ and $\widehat{f}_{i}$. It is designed to compute a cross-modal activation map between visual patch features and text features $g^{i}_{n}$, improving sensitivity to unexpected abnormal patch patterns. Specifically, we design the new encoded feature $f^{*}_{i}$ with a cross-modal normality control coefficient as follows:
\begin{equation}\label{teacher multi-modal feature}
    f^{*}_{i}= f_{i}\oplus\lambda\Psi (\alpha_{i},\beta_{i}),
\end{equation}  
where $f_{i},f^{*}_{i}\in\mathbb{R}^{H\times W\times C}$, $\lambda=1/\|f_{i}\|$ is a scaled coefficient, $\|\cdot\|$ is the $\mathbf{L}_{2}$ norm, and control coefficient $\Psi (\alpha,\beta)$ is written as
\begin{equation}\label{controlitem}
\Psi(\alpha_{i},\beta_{i})=\frac{1}{2}(1+\mathbf{tanh}(\alpha_{i} - \beta_{i})),
\end{equation}
where weight maps $\alpha$ and $\beta$ are defined as:
\begin{equation}\label{apha beta}
\alpha_{i}=f_{i}\otimes g^{i}_{n}, \beta_{i}=f_{i}\otimes g^{i}_{a},
\end{equation}
where $g^{i}_{n}\in\mathbb{R}^{1\times 1\times C}$, $\alpha_{i},\beta_{i}\in \mathbb{R}^{H\times W}$, $\otimes$ denotes the vector-wise product between $g^{i}_{n}$ and each patch embedding $\mathbf{z}_{i}^{l}$ of $f_{i}$, $l$ is the index of patch embedding. Therefore, $\Psi(\alpha_{i},\beta_{i})\in \mathbb{R}^{H\times W}$, and $\oplus$ is element-wise addition operation. Similarly, following Eq.~\eqref{teacher multi-modal feature} and Eq.~\eqref{controlitem}, we also obtain the decoded feature $\widehat{f}_{i}^{*}$ with cross-modal control coefficient:
\begin{equation}\label{student multi-modal feature}
    \widehat{f}^{*}_{i}=\widehat{f}_{i}\oplus\lambda\Psi (\widehat{\alpha}_{i},\widehat{\beta}_{i}),
\end{equation} 
where $\widehat{\alpha}_{i}=\widehat{f_{i}}\otimes g^{i}_{n}$, $\widehat{\beta}_{i}=\widehat{f_{i}}\otimes g^{i}_{a}$. 

We call the above step `feature-level normality promotion' (FNP), as shown in Fig.~\ref{Methodology}. Next, we give a new cross-modal distillation loss to ensure the consistency of the encoded and decoded features with the corresponding control coefficient as follows:
\begin{equation}\label{muti-modal distillation}
    \mathcal{L}_{\mathrm{distill}}=\sum_{i=1}^{3}\Big(1-\frac{\mathbf{Flat }(f^{*}_{i})\cdot\mathbf{Flat }(\widehat{f}_{i}^{*})}{\|\mathbf{Flat }(f^{*}_{i})\|\|\mathbf{Flat }(\widehat{f}_{i}^{*})\|}\Big),
\end{equation}
where $\mathbf{Flat}(\cdot)$ is the fatten function.

\subsubsection{Feature Decoding with Normality Constraint}
In this section, we apply textual features $\{[g^{i}_{n},g^{i}_{a}]\}^{3}_{i=1}$ trained by~\eqref{constrain 1} as anchors to guide feature decoding, alleviating unexpected `OG'. Our solution is to constrain the textual representation of the decoded features not to deviate from ‘normal’ during training, i.e., we also keep class tokens of decoded features aligning with normal text features $\{g^{i}_{n}\}$:
\begin{equation}\label{constrain 2} \mathcal{L}^{2}_{\mathrm{c}}=\sum_{i=1}^{3} -\mathrm{log}\frac{\mathrm{exp}(\widehat{\mathbf{e}}_{i}\cdot g^{i}_{n}/\tau)}{\mathrm{exp}(\widehat{\mathbf{e}}_{i}\cdot g^{i}_{n}/\tau)+\mathrm{exp}(\widehat{\mathbf{e}}_{i}\cdot g^{i}_{a}/\tau)},
\end{equation}
where $\widehat{\mathbf{e}}_{i}$ represents the global feature of $\widehat{f_{i}}$. We combine formulas~\eqref{constrain 1} and~\eqref{constrain 2} to give the cross-modal constraint loss:
\begin{equation}\label{constraint loss}
\mathcal{L}_{\mathrm{constraint}}=\begin{cases}\mathcal{L}^{1}_{\mathrm{c}}\quad\text{if}~~ \text{epoch}<\vartheta 
\\
\mathcal{L}^{1}_{\mathrm{c}}+\gamma\mathcal{L}^{2}_{\mathrm{c}}\quad\text{if}~~\text{epoch}\ge \vartheta
    \end{cases}, 
\end{equation}
where $\gamma=0.1$ is a hyperparameter. Each text feature $g^{i}_{n}$ is a dynamic anchor that is used as a medium to keep the decoded feature toward the ‘normal’ direction.

\subsection{Gated Mixture-of-Experts Module}
\subsubsection{Multi-Layer Fusion}
Following works~\cite{deng2022anomaly,tien2023revisiting}, different layer features of pre-trained encoder are aggregated, improving detection performance. For an input $\mathbf{x}$, we first apply encoder $\mathcal{F}_{\mathrm{E}}$ to extract multi-layer features $\{f_{i}\}^{3}_{i=1}$, and concatenate them as $[f_{1}, f_{2}, f_{3}]$. We adopt a projection layer $\Phi(\cdot)$ (a linear layer with dropout block) to transfer its channel dimension to $C$:
\begin{equation}
    \widetilde{f}=\Phi([f_{1}, f_{2}, f_{3}]),
\end{equation}
where $\widetilde{f}\in \mathbb{R}^{H\times W\times C}$ is a fusion feature (see Fig.~\ref{Methodology} MLF) and is input to the following MoE module.

\subsubsection{Gated Mixture-of-Experts } We employ different expert combination to handle different patch patterns, reducing the mutual interference between them.
Specifically, for a given mini-batch of fusion features, we obtain a batch of patch embedding features $\{\mathbf{z}_{r}\}^{R}_{r=1}$, where $R=B\ast H\ast W$. Our goal is to assign different expert combinations to recognize different patch patterns. Therefore, we first use a router network $\mathbf{G}(\cdot)$ to assign an expert-correlation score to each patch, i.e.,
\begin{equation}\label{gate score}
    \mathbf{H}_{t}=\mathbf{G}_{t}(\mathbf{z}_{r}),~~t\in\{1,2,\cdots, T\},
\end{equation}
where $\mathbf{G}(\cdot): \mathbb{R}^{C}\longmapsto \mathbb{R}^{T}$, $T$ is the number of expert networks $\{\mathbf{E}_{t}(\cdot)\}^{T}_{t=1}$, and each $\mathbf{E}_{i}(\cdot)$ is conducted by a MLP. Next, we select experts with top $K$ scores $\{\mathbf{H}_{k}\}^{K}_{k=1}$ to handle the patch embedding vector $\mathbf{z}_{r}$, denoted by $\{\mathbf{E}_{k}(\cdot)\}^{K}_{k=1}$. Next, we obtain a unique combination of processes on each patch embedding, i.e.,
\begin{equation}\label{moe output}
    \mathbf{z}^{*}_{r}=\sum^{K}_{k=1}\mathbf{H}_{k}*\mathbf{E}_{k}(\mathbf{z}_{r}).
\end{equation}
To prevent the router from assigning dominantly large weights to a few experts, which can lead to a singular scoring operation, we apply a universal importance loss~\cite{bengio2015conditional} to optimize the MoE:
\begin{equation}\label{importance}
    \mathcal{L}_{\mathrm{moe}}=\frac{\mathbf{SD}(\sum^{R}_{r=1}\mathbf{G}(\mathbf{z}_{r}))^{2}}{(\frac{1}{R}\sum^{R}_{r=1}\mathbf{G}(\mathbf{z}_{r}))^{2}+\varepsilon},
\end{equation}
where $\mathbf{SD}(\cdot)$ is standard deviation operation, $\varepsilon$ is added for numerical stability. Finally, according to Eq.~\eqref{muti-modal distillation}, ~\eqref{constraint loss}, and~\eqref{importance}, we obtain a total loss to train our model:
\begin{equation}
    \mathcal{L}_{\mathrm{total}}=\mathcal{L}_{\mathrm{distill}}+\mathcal{L}_{\mathrm{constraint}}+\mathcal{L}_{\mathrm{moe}}.
\end{equation}

\subsection{Inference}
The inference is consistent with the training phase. We apply encoder $\mathcal{F}_{\mathrm{E}}(\cdot)$, learned prompts $\{[\mathbf{p}^{i}_{n},\mathbf{p}^{i}_{a}]\}^{3}_{i=1}$, multi-layer fusion $\Phi(\cdot)$, MoE module, and decoder $\mathcal{F}_{\mathrm{D}}(\cdot)$ to produce encoded features $\{f^{*}_{i}\}^{3}_{i=1}$ and decoded features $\{\widehat{f}_{i}^{*}\}^{3}_{i=1}$.  We design a pixel-level anomaly score map as:
\begin{equation}\label{anomaly score}
    \mathcal{S}(f^{*}_{i},\widehat{f}_{i}^{*})=\sum_{i=1}^{3}\sigma_{i}\Big(1-\mathbf{d}(f^{*}_{i},\widehat{f}_{i}^{*})\Big),
\end{equation}
where $\mathbf{d}(\cdot,\cdot)$ is pixel-wise cosine similarity, $\sigma(\cdot)$ is the upsampling factor in order to keep
the same size as the input image. In addition, the image-level anomaly score is defined as the maximum score of the pixel-level anomaly map.
\section{Experiments}
\subsection{Experimental Setup}

\begin{table*}[t]
\centering
\small
\selectfont{\begin{tabular}{ccccc}
\toprule
Method $\longrightarrow$     & \textbf{RD4AD*}~\cite{deng2022anomaly}                    & \textbf{UniAD*}~\cite{you2022unified}                 & \textbf{DiAD}~\cite{he2024diffusion}                    & \textbf{CND}                    \\ \cline{1-1}
Category $\downarrow $  & \textbf{CVPR2022}                 & \textbf{NeurIPS2022}              & \textbf{AAAI2024}                 & \textbf{Ours}                     \\ \hline
Bottle     & 99.6/96.1/91.1/99.9/48.6 & 97.0/98.1/\underline{93.1}/\textbf{100.}/\underline{66.0} & \underline{99.7}/\underline{98.4}/86.6/99.5/52.2 & \textbf{100.}/\textbf{99.0}/\textbf{97.1}/\textbf{100.}/\textbf{81.8} \\

Cable      & 84.1/85.1/75.1/89.5/26.3 & \underline{95.2}/\underline{97.3}/\underline{86.1}/95.9/39.9  & 94.8/96.8/80.5/\underline{98.8}/\underline{50.1} & \textbf{98.9}/\textbf{98.2}/\textbf{92.5}/\textbf{99.3}/\textbf{64.1} \\
Capsule    & \underline{94.1}/\textbf{98.8}/\textbf{94.8}/96.9/\textbf{43.4} & 86.9/\underline{98.5}/92.1/97.8/\underline{42.7} & 89.0/97.1/87.2/97.5/42.0 & \textbf{98.0}/98.2/\underline{93.8}/\textbf{99.3}/36.9 \\

Hazelnut   & 60.8/97.9/92.7/69.8/36.2 & \underline{99.8}/98.1/\underline{94.1}/\textbf{100.}/\underline{55.2} & 99.5/\underline{98.3}/91.5/99.7/\textbf{79.2} & \textbf{100.}/\textbf{98.8}/\textbf{94.9}/\textbf{100.}/53.3 \\
Metal Nut  & \textbf{100.}/94.8/\textbf{91.9}/\textbf{100.}/\underline{55.5} & 99.2/62.7/81.8/99.9/14.6  & 99.1/\textbf{97.3}/\underline{90.6}/96.0/30.0 & \textbf{100.}/\underline{95.5}/89.2/\textbf{100.}/\textbf{68.4 }\\ 
Pill       & \textbf{97.5}/\underline{97.5}/\underline{95.8}/\textbf{99.6}/\underline{63.4} & 93.7/95.0/95.3/98.7/44.0  & 95.7/95.7/89.0/98.5/46.0 & \underline{96.8}/\textbf{98.8}/\textbf{96.1}/\underline{99.5}/\textbf{80.2} \\
Screw      & \textbf{97.7}/\textbf{99.4}/\textbf{96.8}/\underline{99.3}/\underline{40.2} & 87.5/98.3/\underline{95.2}/96.5/28.7  & 90.7/97.9/95.0/\textbf{99.7}/\textbf{60.6} & \underline{93.8}/\underline{99.0}/94.7/98.3/26.2 \\ 
Toothbrush & 97.2/\textbf{99.0}/92.0/99.0/\underline{53.6} & 94.2/98.4/87.9/97.4/34.9  & \textbf{99.7}/\textbf{99.0}/\textbf{95.0}/\textbf{99.9}/\textbf{78.7} & \underline{99.5}/\textbf{99.0}/\underline{93.0}/\underline{99.8}/49.7 \\
Transistor & 94.2/85.9/74.7/95.2/42.3 & \textbf{99.8}/\textbf{97.9}/\textbf{93.5}/98.0/\textbf{59.5}  & \textbf{99.8}/\underline{95.1}/\underline{90.0}/\textbf{99.0}/15.6 & 96.0/94.5/74.1/94.4/57.3 \\ 
Zipper     & \textbf{99.5}/\textbf{98.5}/\textbf{94.1}/\textbf{99.9}/\underline{53.9} & 95.8/96.8/92.6/99.5/40.1  & \underline{99.3}/96.2/91.6/99.8/\textbf{60.7} & 99.1/\underline{97.6}/\underline{93.8}/\underline{99.8}/45.1 \\
Carpet     & 98.5/\underline{99.0}/\underline{95.1}/99.6/\underline{58.5} & \underline{99.8}/98.5/94.4/\textbf{99.9}/49.9 & 99.4/98.6/90.6/\textbf{99.9}/42.2 & \textbf{99.9}/\textbf{99.3}/\textbf{97.0}/\textbf{99.9}/\textbf{70.7} \\
Grid       & 98.0/\underline{96.5}/\textbf{97.0}/99.4/23.0 & 98.2/63.1/92.9/99.5/10.7  & \underline{98.5}/96.6/94.0/\textbf{99.8}/\textbf{66.0} & \textbf{99.1}/\textbf{98.4}/\underline{95.0}/\underline{99.7}/\underline{25.8} \\
Leather    & \textbf{100.}/\underline{99.3}/\underline{97.4}/\textbf{100.}/38.0 & \textbf{100.}/98.8/96.8/\textbf{100.}/32.9  & 99.8/98.8/91.3/99.7/\textbf{56.1} & \textbf{100.}/\textbf{99.5}/\textbf{98.7}/\textbf{100.}/\underline{50.1} \\
Tile       & 98.3/\underline{95.3}/85.8/99.3/48.5 & \underline{99.3}/91.8/78.4/99.8/42.1  & 96.8/92.4/\underline{90.7}/\underline{99.9}/\underline{65.7} & \textbf{100.}/\textbf{97.7}/\textbf{94.1}/\textbf{100.}/\textbf{73.4} \\
Wood       & \underline{99.2}/\underline{95.3}/90.0/\underline{99.8}/\underline{47.8} & 98.6/93.2/86.7/99.6/37.2  & \textbf{99.7}/93.3/\textbf{97.5}/\textbf{100.}/43.3 & 98.3/\textbf{96.4}/\underline{91.4}/99.5/\textbf{63.3} \\ \hline
Mean       & 94.6/96.1/\underline{91.1}/96.5/48.6 & 96.5/96.8/90.7/98.8/43.4  & \underline{97.2}/96.8/90.7/\underline{99.0}/\underline{52.6} & \textbf{98.6}/\textbf{98.0}/\textbf{93.0}/\textbf{99.3}/\textbf{56.4} \\ \hline
mTotal     & 85.4                     & 85.2                                          & \underline{87.3}                     & \textbf{89.0}                     \\ \bottomrule
\end{tabular}}
\caption{Comprehensive anomaly detection results with image-level AUROC, pixel-level AUROC, AUPRO, image-level mAP, and pixel-level mAP metrics on \textbf{MVTec AD} dataset. We also provide the average of all metrics at the bottom of the table. The best and second-best results are in bold and underlined, respectively. *: The results are sourced from~\cite{he2024mambaad}.} 
\label{MVTec performance}
\end{table*}

\subsubsection{Datasets}
MVTec AD~\cite{bergmann2019mvtec} is the most widely used industrial anomaly detection dataset, containing $15$
categories of sub-datasets. The training set consists of $3629$ images with anomaly-free samples. The test dataset includes $1725$ normal and abnormal images. Segmentation masks are provided for anomaly localization evaluation.
VisA~\cite{zou2022spot} is a challenging AD dataset containing $10821$ images and $12$ categories of sub-datasets.

\subsubsection{Evaluation Metrics}
Following the prior work~\cite{he2024diffusion}, image-level Area Under the
Receiver Operating Characteristic Curve (I-AUROC) and Average Precision (I-mAP) are applied for anomaly classification. Pixel-level AUROC, pixel-level mAP, and  AUPRO~\cite{bergmann2020uninformed} are used for anomaly localization.

\subsubsection{Implementation Details}
The implementation is based on
Pytorch. The publicly
available CLIP model (VIT-L/14@336px) is the backbone of our method. We select the Adam optimizer to train our model. Then, we resize the resolution of each image to $224\times 224$. In addition, the length of each learnable text prompt is set to $12$, consistent with previous work \cite{zhou2023anomalyclip}. For both datasets, we set temperature coefficient $\tau=0.001$ and batch size to $8$ with learning rate $0.001$ to train the whole model. The number of experts is set to $5$ with top $K=2$ gated scores in the MoE. Next, we set the epoch to $250$ and $200$ for MVTec AD and VisA with the same $\vartheta=5$, respectively. All experiments are conducted on a single
NVIDIA Tesla V100 32GB GPU.
\subsection{Comprehensive Comparisons with SOTA Methods}
In this section, we compare our approach with several SOTA methods on MVTec AD and VisA datasets, where $5$ different metrics, I-AUROC, P-AURPOC, AUPRO, I-mAP, P-mAP, are shown for comprehensive evaluation in Table~\ref{MVTec performance} and Table~\ref{tab:visa}, respectively. 
\subsubsection{Results on MVTec AD}
\begin{figure}[t]
\centering
\includegraphics[width=\columnwidth]{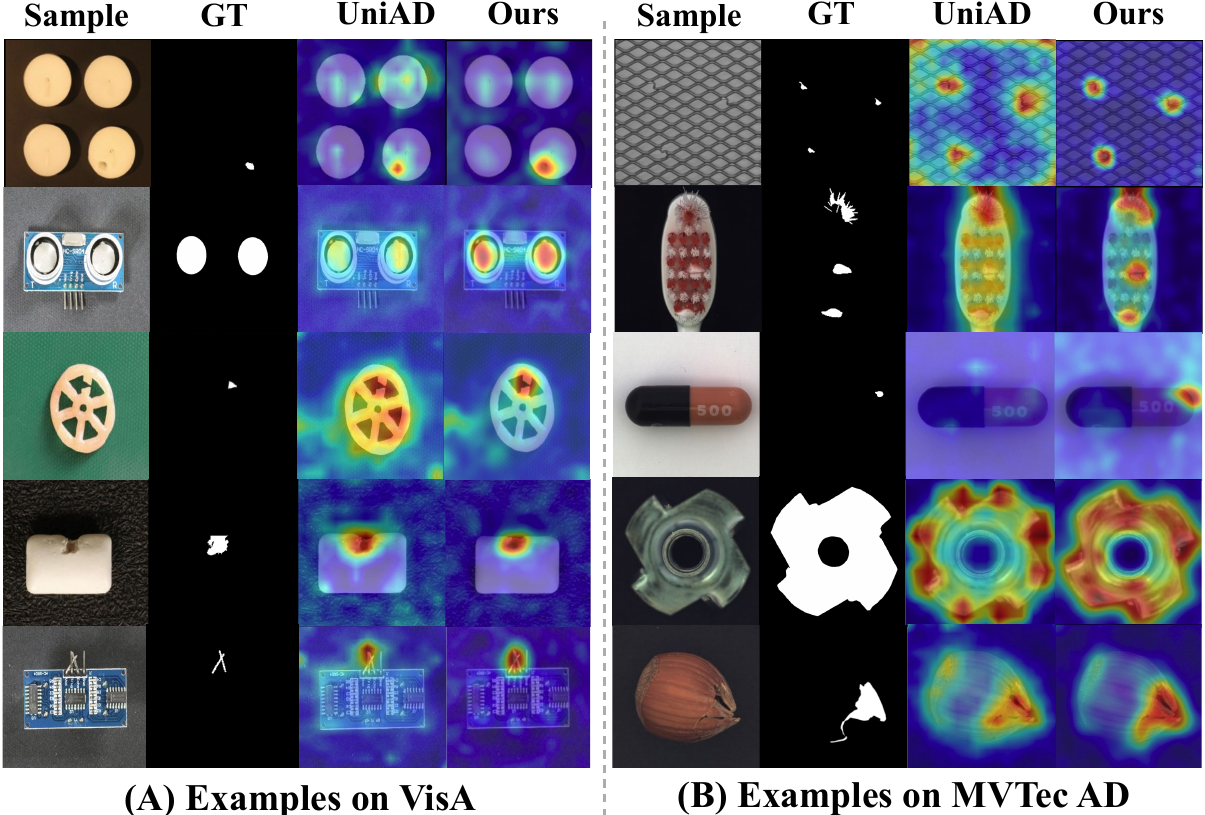}
\caption{Visualization for detection results of UniAD and our method on MVTec AD and VisA datasets.}
\label{fig: visualization }
\end{figure}

As reported in Table~\ref{MVTec performance}, for the widely used MVTec AD dataset, our cross-modal normality distillation framework (CND) achieves five different metrics by $98.6/98.0/93.0/99.3/56.4$, and the mean performance of five metrics by $89.0$ under multi-class setting. Compared to UniAD and DiAD, our method improves five metrics by $+2.1/1.2/2.3/0.5/13.0$ and $+1.4/1.2/2.3/0.3/3.8$, and the mean metric by $+3.8$ and $+1.7$, respectively. Our method significantly outperforms the single-modal distillation framework, RD4AD, by $+4.0/1.9/1.9/2.8/7.8$, in terms of five metrics. In addition, our method is more stable than previous methods, achieving a performance of $93.8+$ in terms of image and pixel-level AUROC for all categories. However, RD4AD merely achieves a $60.8$ I-AUROC metric on Hazelnut, and UniAD achieves a $63.1$ P-AUROC metric on Grid. To further illustrate the effectiveness
of our proposed method in anomaly localization, we visualize UniAD and our method prediction detection results in Fig.~\ref{fig: visualization } (B).

\subsubsection{Results on VisA}
As reported in Table~\ref{tab:visa}, for the VisA dataset, our proposed method also achieves a SOTA performance, namely 93.2 and 98.5 in terms of I-AUROC and P-AUROC metrics, respectively. Compared to previous multi-class methods, our method outperforms UniAD by $+7.7/2.6/15.8/7.1/16.8$, and DiAD by $+6.4/2.5/16.2/4.3/11.7$. Especially, in the `fryum' and `capsules' categories, our method greatly improves anomaly classification compared to UniAD~\cite{you2022unified} and \cite{he2024diffusion}. Finally, we also visualize our obtained localization result via heat maps on VisA in Fig.~\ref{fig: visualization } (A).

\subsection{Ablation Study}
In this section, we investigate the contribution of different main
components in our approach. Additionally, we show results on different backbones with different resolutions and investigate the impact of hyperparameters of MoE.
\begin{table}[t]
\centering
\selectfont{
\begin{tabular}{cccccc}
\toprule
               & MLF        & CNC       & MoE       & Performance        & Mean \\ \midrule
\romannumeral1 & \ding{55} & \ding{55} &\ding{55} & 95.4/95.6/90.3 & 93.7 \\ 
\romannumeral2 & \ding{51} & \ding{55} & \ding{55} & 96.0/96.3/91.1 & 94.4 \\
\romannumeral3 & \ding{55} & \ding{51} & \ding{55} & 96.7/97.0/92.0 & 95.2 \\
\romannumeral4 &\ding{55} & \ding{55} & \ding{51} & 96.3/95.8/90.7 & 94.2 \\
\romannumeral5 & \ding{51} & \ding{51} & \ding{55} & \underline{98.1}/\underline{97.7}/\underline{92.8} & \underline{96.2} \\
\romannumeral6 & \ding{51} & \ding{55} & \ding{51} & 97.0/96.8/91.4 & 95.0 \\\midrule
\romannumeral7 & \ding{51} & \ding{51} & \ding{51} & \textbf{98.6}/\textbf{98.0}/\textbf{93.0} & \textbf{96.5} \\ \bottomrule
\end{tabular}}
\caption{Ablation study of our method on MVTec AD. MLF: Multi-Layer Fusion. CNC: Cross-modal Normality Constraint, including feature-level normality promotion, constraint loss $\mathcal{L}_{\mathrm{constraint}}$, and distillation loss $\mathcal{L}_{\mathrm{distill}}$. MoE: Gated Mixture-of-Expert Module. \textbf{Bold}/\underline{underline} values indicate the best/runner-up.}
\label{tab:ablation}
\end{table}

\begin{figure}[t]
\centering
\includegraphics[width=\columnwidth]{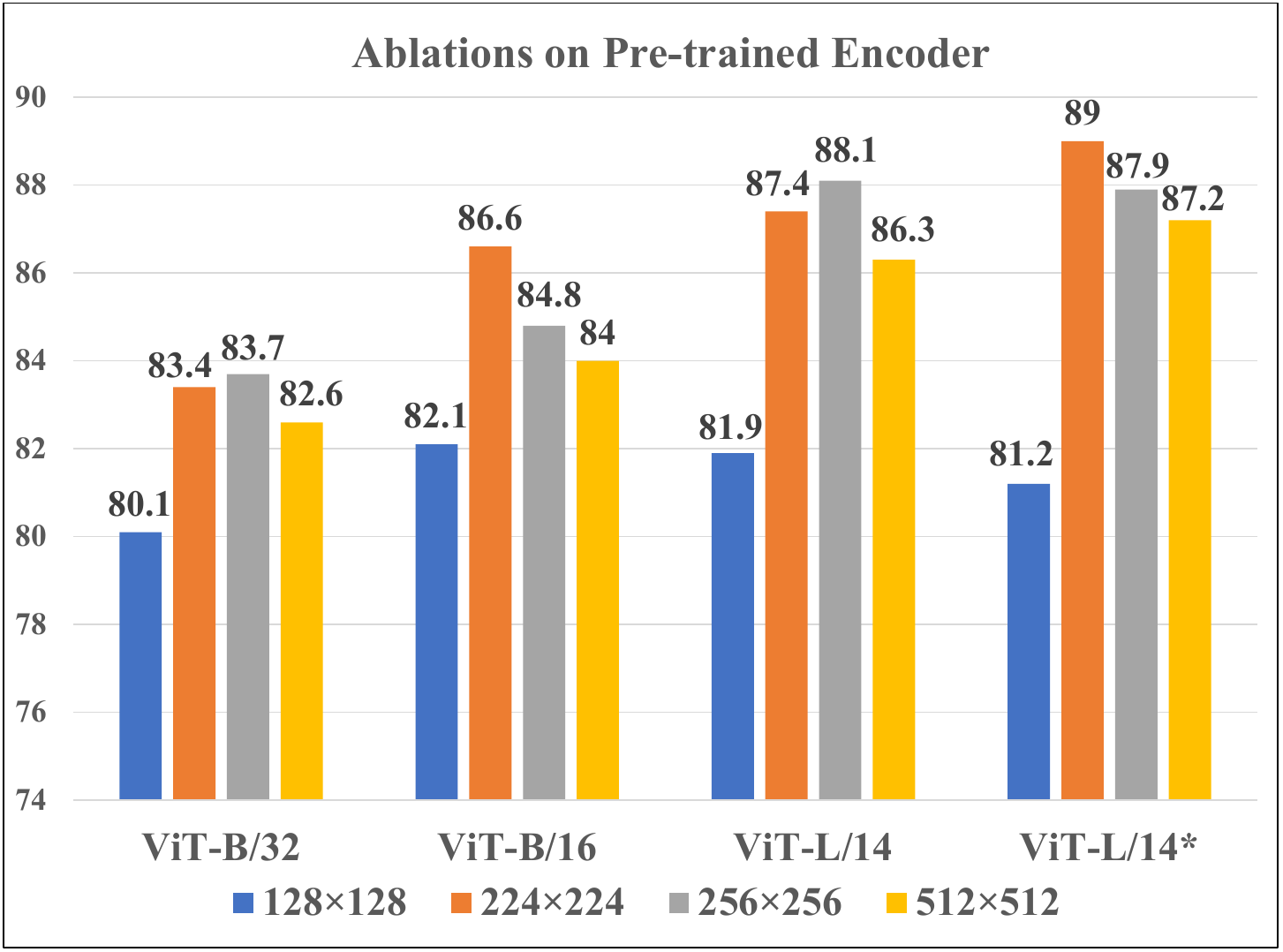}
\caption{Choices on four pre-trained teacher network (encoder) with four different resolutions. The vertical axis represents the average value of I-AUROC/P-AUROC/AUPRO/I-mAP/P-mAP. ViT-L/14* denotes pre-trained CLIP model, VIT-L/14@336px.}
\label{fig: ablations of backbone }
\end{figure}

\begin{table*}[t]
\centering
\small
\selectfont{\begin{tabular}{cccccc}
\toprule
Method $\longrightarrow$    & \textbf{RD4AD*}~\cite{deng2022anomaly}    & \textbf{UniAD*}~\cite{you2022unified}     & \textbf{DiAD*}~\cite{he2024diffusion}                     & \textbf{CND} \\ \cline{1-1}
Category $\downarrow $  & \textbf{CVPR2022}                 & \textbf{NeurIPS2022}              & \textbf{AAAI2024}                 & \textbf{Ours} \\ \hline
pcb1       & \textbf{96.2}/\underline{99.4}/\textbf{95.8}/\textbf{95.5}/\underline{66.2} & 92.8/93.3/64.1/\underline{92.7}/~~3.9 & 88.1/98.7/80.2/88.7/49.6 & \underline{94.1}/\textbf{99.5}/\underline{92.6}/91.7/\textbf{70.7} \\
pcb2       & \textbf{97.8}/\underline{98.0}/\textbf{90.8}/\textbf{97.8}/\textbf{22.3} & 87.8/93.9/66.9/87.7/~~4.2  & 91.4/95.2/67.0/91.4/~~7.5  & \underline{95.9}/\textbf{98.4}/\underline{88.8}/\underline{92.2}/\underline{18.1} \\
pcb3       & \textbf{96.4}/\underline{97.9}/\textbf{93.9}/\textbf{96.2}/\textbf{26.2} & 78.6/97.3/70.6/78.6/13.8 & 86.2/96.7/68.9/87.6/~~8.0  & \underline{92.0}/\textbf{98.6}/\underline{93.7}/\underline{93.9}/ \underline{21.7}\\
pcb4       & \textbf{99.9}/\underline{97.8}/\underline{88.7}/\textbf{99.9}/\underline{31.4} & 98.8/94.9/72.3/98.8/14.7 & 99.6/97.0/85.0/99.5/17.6 & \textbf{99.9}/\textbf{99.0}/\textbf{90.5}/\underline{99.8}/ \textbf{40.5}\\
macaroni1  & 75.9/\textbf{99.4}/\textbf{95.3}/61.5/~~2.9  & 79.9/97.4/84.0/79.8/~~3.7   & 85.7/94.1/68.5/\textbf{85.2}/\textbf{10.2} & \textbf{86.7}/\underline{98.6}/\underline{90.5}/\underline{84.4}/~~\underline{7.8} \\
macaroni2   & \textbf{88.3}/\textbf{99.7}/\textbf{97.4}/\textbf{84.5}/\textbf{13.2} & 71.6/95.2/76.6/71.6/~~0.9    & 62.5/93.6/73.1/57.4/~~0.9  & \underline{84.4}/\underline{98.1}/\underline{93.6}/\underline{81.3}/\underline{12.7} \\
capsules   & \underline{82.2}/\textbf{99.4}/\textbf{93.1}/\textbf{90.4}/\textbf{60.4} & 55.6/88.7/43.7/55.6/~~3.0   & 58.2/97.3/77.9/69.0/10.0 & \textbf{83.4}/\underline{98.4}/\underline{88.
6}/\underline{89.2}/\underline{33.6} \\ 
candle     & 92.3\textbf{/99.1}/\textbf{94.9}/\underline{92.9}/\textbf{25.3} & \textbf{94.1}/\underline{98.5}/91.6/\textbf{94.0}/\underline{17.6}  & 92.8/97.3/89.4/92.0/12.8 & \underline{93.7}/98.4/\underline{91.9}/90.0/16.7 \\
cashew     & 92.0/91.7/86.2/\textbf{95.8}/44.2 & \underline{92.8}/\textbf{98.6}/\textbf{87.9}/92.8/51.7 & 91.5/90.9/61.8/\underline{95.7}/53.1 & \textbf{94.1}/\underline{98.1}/\underline{87.4}/92.8/\underline{62.9} \\
chewinggum & 94.9/98.7/76.9/97.5/\underline{59.9} & 96.3/\underline{98.8}/\underline{81.3}/96.2/54.9 & \textbf{99.1}/94.7/59.5/\textbf{99.5}/11.9 & \underline{98.7}/\textbf{99.1}/\textbf{89.4}/\underline{99.2}/\textbf{61.3}\\
fryum      & \underline{95.3}/\underline{97.0}/\textbf{93.4}/\textbf{97.9}/\underline{47.6} & 83.0/95.9/76.2/83.0/34.0& 89.8/\textbf{97.6}/81.3/95.0/\textbf{58.6} & \textbf{96.4}/\underline{97.0}/\underline{92.1}/\textbf{97.9}/47.3\\ 
pipe-fryum & \underline{97.9}/\underline{99.1}/\underline{95.4}/\underline{98.9}/56.8 & 94.7/98.9/91.5/94.7/50.2  & 96.2/\textbf{99.4}/89.9/98.1/\textbf{72.7} & \textbf{98.9}/98.4/\textbf{97.5}/\textbf{99.0}/\underline{61.4} \\ \hline
Mean       & \underline{92.4}/\underline{98.1}/\textbf{91.8}/\underline{92.4}/\textbf{38.0} & 85.5/95.9/75.6/85.5/21.0 & 86.8/96.0/75.2/88.3/26.1 & \textbf{93.2}/\textbf{98.5}/\underline{91.4}/\textbf{92.6}/\underline{37.8}                        \\ \hline
mTotal     & \underline{82.5}                     &  72.7                                         &   74.5                   &   \textbf{82.7}                   \\ \bottomrule
\end{tabular}}
\caption{Comprehensive anomaly detection results with five different metrics on \textbf{VisA} dataset. \textbf{Bold}/\underline{underline} values indicate the best/runner-up. *: The results are sourced from~\cite{he2024mambaad}.} 
\label{tab:visa}
\end{table*}

\begin{table}[t]
\centering
\small
\setlength\tabcolsep{5pt}
\selectfont{
\begin{tabular}{ccccc}
\toprule
\textbf{Top} \textbf{K} $\longrightarrow $ &                       &                       &                       &                       \\ \cline{1-1}
\textbf{No.Experts} $\downarrow$ & \multirow{-2}{*}{$K=1$} & \multirow{-2}{*}{$K=2$} & \multirow{-2}{*}{$K=3$} & \multirow{-2}{*}{$K=4$} \\ \hline
None                  & 98.1/97.7             & \ding{55}                   & \ding{55}                    & \ding{55}                    \\ 
$T=1$                     & 98.0/97.7            & \ding{55}             & \ding{55}                    & \ding{55}                   \\
$T=2$                     & 97.4/96.7             & 97.2/96.9             & \ding{55}                   & \ding{55}                   \\
$T=3$                     & 97.4/97.0             & 98.0/97.3             & 97.5/97.5             & \ding{55}                    \\
$T=4$                     & 97.7/97.2             & 98.2/97.4            & 97.9/\underline{97.8}           & 97.3/97.6             \\
$T=5$                     & 97.6/97.4             & \textbf{98.6}/\textbf{98.0}             & 98.2/97.7            & 98.0/97.7            \\ 
$T=6$                     & 97.3/97.3             & \underline{98.5}/97.3             & 97.8/97.7             & 97.4/97.4             \\
\bottomrule
\end{tabular}}
\caption{Impact of Hyperparameters in MoE module, where I-AUROC and P-AUROC metrics are reported on MVTec AD dataset. $T$ and $K$ denote the number of experts and Top $K$ coefficient respectively, and $K\le T$.
\textbf{Bold}/\underline{underline} values indicate the best/runner-up.}
\label{tab:moe}
\end{table}

\subsubsection{Effectiveness of Main Components}
In Table \ref{tab:ablation}, we report three main metrics, including I-AUROC, P-AUROC and AUPRO, to study the impact of each key component on MVTec AD.  As shown in line {\romannumeral1} and {\romannumeral3} of Table \ref{tab:ablation}, CNC improves our base model by $+1.3/1.4/1.7$. In addition, equipped with `MLF' module, we get a gain of $+2.1/1.4/1.7$ via CNC (see {\romannumeral2} and {\romannumeral5} in Table \ref{tab:ablation}). Therefore, our proposed CNC successfully suppresses undesired `OG'. 
In addition, the proposed MoE module alleviates `OG' by assigning different weights to different patch patterns, improving performance by $+1.0/0.5$ in terms of I-AUROC and P-AUROC (see {\romannumeral2} and {\romannumeral6}).
We also found that MoE  enhances image-level anomaly detection, enhancing $+0.5$ on I-AUROC metric (see {\romannumeral5} and {\romannumeral7} in Table \ref{tab:ablation}). Finally, our designed multi-layer fusion module can well fuse different layer information of ViT. As shown {\romannumeral1} and {\romannumeral2} in Table \ref{tab:ablation}, MLF improves three metrics by $+0.6/0.7/0.8$. 

\subsubsection{Choices on Pre-trained Encoders and Image Resolutions}
Fig.~\ref{fig: ablations of backbone } shows results of different resolutions for four pre-trained encoder. On the one hand, we found that models that perform well in zero-shot classification have higher performance in our framework. We obtain the highest performance of $89.0$ when applying ViT-L/14*, but the performance rapidly degraded when using ViT-B/32 and ViT-B/16. On the other hand, we found that both low and high resolutions ($128\times128$ and $512\times512$) degrade detection performance and great performances can be achieved with resolutions $224\times224$ and $256\times256$, which is consistent with previous work~\cite{deng2022anomaly}.

\subsubsection{Impact of Hyperparameters in MoE}
According to Table~\ref{tab:moe}, an appropriate selection on hyperparameter greatly improves anomaly localization and classification for our method. When $T=1$, it is equivalent to connecting an adapter and does not significantly affect performance. Both large and small values of $T$ can degrade performance. We consider that large $T$ may lead to some experts under-fitting and small $T$ may result in some experts over-fitting. when $T=5$ and $K=2$, the best performance is achieved.

\section{Conclusion}
In this paper, we propose a cross-modal distillation framework to address the inevitable `over-generalization' in multi-class training. Firstly, we propose cross-modal normality constraint (CNC) to guide decoded features to align the decoded features with a textual representation of normality, thereby improving the normality of the distilled features and final detection performance. We also propose a gated MoE module to re-weight different patch patterns, reducing the mutual interference between them. Finally, extensive experiments show that our method achieves competitive performance on MVTec AD and VisA datasets. 

\section{Acknowledgments}
This work was supported by the National Natural Science Foundation of China (No. 62471405, 62331003, 62301451), Jiangsu Basic Research Program Natural Science Foundation (SBK2024021981), Suzhou Basic Research Program (SYG202316) and XJTLU REF-22-01-010, XJTLU AI University Research Centre, Jiangsu Province Engineering Research Centre of Data Science and Cognitive Computation at XJTLU and SIP AI innovation platform (YZCXPT2022103).

\bibliography{aaai25}

\end{document}